\documentclass[10pt,twocolumn,letterpaper]{article}

\usepackage[pagenumbers]{cvpr} % To force page numbers, e.g. for an arXiv version

\usepackage{multirow}
\usepackage[normalem]{ulem}
\useunder{\uline}{\ul}{}
\usepackage[dvipsnames, svgnames, x11names]{xcolor}% 颜色支持
\usepackage{colortbl}
\usepackage[dvipsnames]{xcolor}

\definecolor{cvprblue}{rgb}{0.21,0.49,0.74}
\usepackage[pagebackref,breaklinks,colorlinks,citecolor=cvprblue]{hyperref}

\def\paperID{249} % *** Enter the Paper ID here
\def\confName{3DV\xspace}
\def\confYear{2024\xspace}

\title{MC-Stereo: Multi-Peak Lookup and Cascade Search Range for Stereo Matching}
\author{Shiyu Zhao\textsuperscript{1,}\thanks{Correspondence
to: Shiyu Zhao (sz553@rutgers.edu).} \qquad 
Long Zhao\textsuperscript{2}  \qquad 
Zhixing Zhang\textsuperscript{1}  \qquad 
Enyu Zhou\textsuperscript{3} \qquad 
Dimitris Metaxas\textsuperscript{1}\\
\\
\textsuperscript{1}Rutgers University \qquad
\textsuperscript{2}Google Research\qquad
\textsuperscript{3}SenseTime Research\\
}

\author{Miaojie Feng$^{1,}$\footnotemark[1] \quad\quad Junda Cheng$^{1,}$\footnotemark[1] \quad\quad Hao Jia$^{1}$ \quad\quad Longliang Liu$^{1}$ \quad\quad Gangwei Xu$^{1}$ \quad\quad\\
Qingyong Hu$^{2}$ \quad\quad Xin Yang$^{1,}$\footnotemark[2]\\
{$^{1}$School of EIC, Huazhong University of Science and Technology}\\
{$^{2}$Academy of Military Sciences, Beijing, China}
}
\begin{document}
\maketitle
\begin{abstract}
Stereo matching is a fundamental task in scene comprehension. In recent years, the method based on iterative optimization has shown promise in stereo matching. However, the current iteration framework employs a single-peak lookup, which struggles to handle the multi-peak problem effectively. Additionally, the fixed search range used during the iteration process limits the final convergence effects. To address these issues, we present a novel iterative optimization architecture called MC-Stereo. This architecture mitigates the multi-peak distribution problem in matching through the multi-peak lookup strategy, and integrates the coarse-to-fine concept into the iterative framework via the cascade search range. Furthermore, given that feature representation learning is crucial for successful learn-based stereo matching, we introduce a pre-trained network to serve as the feature extractor, enhancing the front end of the stereo matching pipeline. Based on these improvements, MC-Stereo ranks first among all publicly available methods on the KITTI-2012 and KITTI-2015 benchmarks, and also achieves state-of-the-art performance on ETH3D. Code is available at~\href{https://github.com/MiaoJieF/MC-Stereo}{https://github.com/MiaoJieF/MC-Stereo}.
\end{abstract}

{
\renewcommand{\thefootnote}{\fnsymbol{footnote}}
\footnotetext[1]{Authors contributed equally.}
\footnotetext[2]{Corresponding author.}
}

\section{Introduction}
\label{sec:intro}

\hspace{1em} Stereo matching is a crucial computer vision technology. Its primary objective is to establish the corresponding relationship between the identical scene or object in 3D space from two images. This technology finds widespread applications in various fields, including computer vision, robot navigation, augmented reality, virtual reality, and autonomous driving.

%%%%%%%%%%%%%%%%%%%%%%%%%%%%%%%%%%%%%%%%%%%%%%%%%%%%%%%%%%%%%%%%%%%%%%%%
% 代价体可视化
\begin{figure}[t]
    \centering
    \includegraphics[width=0.9\linewidth]{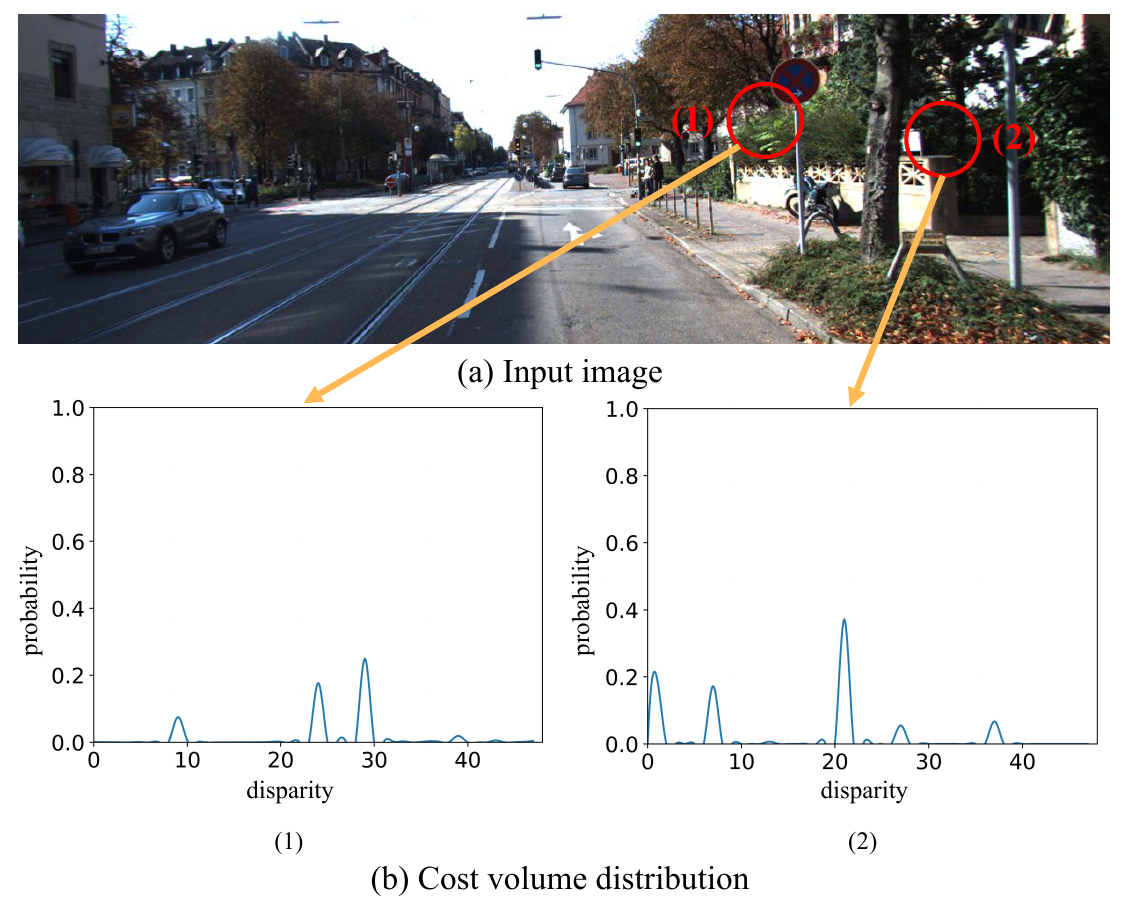}
    \vspace{-10pt}
    \caption{Illustration of multi-peak distribution in the cost volume. (a) Input left image. (b) The cost volume distribution of a single pixel in the center of the red circle.}
    \label{fig:cost_volume_vis}
    \vspace{-15pt}
\end{figure}
%%%%%%%%%%%%%%%%%%%%%%%%%%%%%%%%%%%%%%%%%%%%%%%%%%%%%%%%%%%%%%%%%%%%%%%%

%%%%%%%%%%%%%%%%%%%%%%%%%%%%%%%%%%%%%%%%%%%%%%%%%%%%%%%%%%%%%%%%%%%%%%%%
% KITTI可视化图
\begin{figure*}[t]
    \centering
    \includegraphics[width=0.90\linewidth]{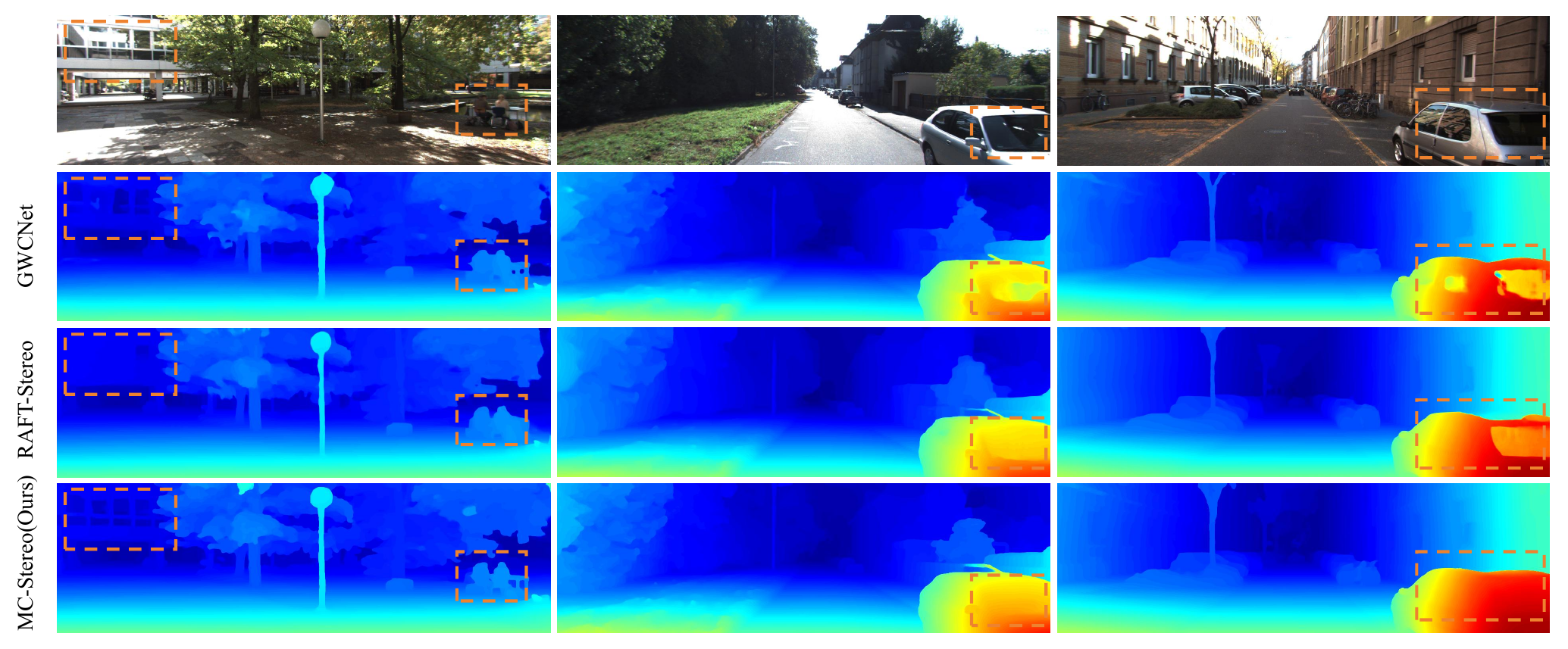}
    \vspace{-5pt}
    \caption{Qualitative results on KITTI. The second, third, and fourth rows are the results of GWCNet~\cite{guo2019group}, RAFT-Stereo~\cite{lipson2021raft} and our MC-Stereo respectively. Our MC-Stereo performs better in reflective areas.}
    \label{fig:vis_kitti}
    \vspace{-10pt}
\end{figure*}
%%%%%%%%%%%%%%%%%%%%%%%%%%%%%%%%%%%%%%%%%%%%%%%%%%%%%%%%%%%%%%%%%%%%%%%%

Traditional stereo matching methods~\cite{zhang2009cross, sun2003stereo, hirschmuller2007stereo} primarily depend on manually-crafted features and matching strategies, which exhibit suitability for simple scenes but may show limitations in handling complex scenes. The deep learning-based stereo matching method~\cite{mayer2016large, chang2018pyramid, guo2019group, xu2022attention} improves the accuracy to a new height. Early deep learning-based stereo matching methods usually consist of 4 steps, including feature extraction, cost volume construction, cost aggregation, and disparity regression. These methods typically employ multi-layer 3D convolution for cost aggregation to regularize the cost volume. In recent years, the method~\cite{teed2020raft, jiang2021learning} based on iterative optimization has achieved milestone progress in optical flow estimation. Subsequently, RAFT-Stereo~\cite{lipson2021raft} introduces the framework of iterative optimization to stereo matching. Specifically, after constructing the cost volume, RAFT-Stereo~\cite{lipson2021raft} iteratively retrieves local costs through a GRU-based updater and regresses the disparity residual. This updater can preserve contextual information and previous states in hidden layers, and thus it can effectively combine context with local matching details and in turn progressively improve disparity estimation accuracy. However, there are two problems with the existing iteration framework~\cite{lipson2021raft,xu2023iterative}. (1) We observe that the cost volume usually contains multi-peak distribution (as shown in Figure~\ref{fig:cost_volume_vis}), and the existing iterative optimization based methods use single-peak lookup, which can not deal with the problem of multi-peak distribution well. (2) The process of iterative optimization should be modeled as a process of gradual refinement. However, the existing framework method uses a single fixed search range, which limits its final convergence effect.

In this paper, we propose the multi-peak lookup strategy and cascade search range to solve these two problems. The purpose of multi-peak lookup strategy is to improve the ability of the model to handle multi-peak distribution. In each iteration, the multi-peak lookup strategy indexes several local cost volumes with high probability, then integrates information through GRU, and finally regresses the disparity. The cascade search range divides the iterative optimization process into multiple stages, using a specific search range at each stage. Specifically, we use a large search range at the beginning of iterative optimization and gradually reduce the search range according to a specific pattern. This design is based on the idea of coarse to fine, a large search range at the beginning is conducive to accelerating convergence, and a small search range at the end is conducive to finer results. Experiments show that the proposed method can effectively improve the performance. In addition, feature representation learning is the key to realizing learning-based stereo matching. However, in the field of stereo matching, feature extraction has not been improved significantly for many years. We introduce a network pre-trained on a large dataset as feature extractor to improve the front end deficiencies in pipeline. We name our method MC-Stereo. Based on these improvements, MC-Stereo ranks $1^{st}$ among all publicly available methods on the KITTI-2012~\cite{geiger2012we} and KITTI-2015~\cite{menze2015joint} benchmarks, and also achieves state-of-the-art performance on ETH3D~\cite{schops2017multi}. Our contribution can be summarized as follows:

\begin{itemize}[leftmargin=*]
    \vspace{0pt}
    \item We propose the multi-peak lookup strategy to improve the ability of the model in dealing with multi-peak distribution.
    \vspace{0pt}
    \item We propose a cascade search strategy, which combines the idea of coarse to fine into the iterative optimization framework.
    \vspace{0pt}
    \item Our proposed MC-Stereo ranks $1^{st}$ among all published methods on the KITTI-2012 and KITTI-2015 benchmarks, and also achieves SOTA performance on ETH3D. 
    \vspace{0pt}
\end{itemize}

\section{Related Work}
\label{sec:related}

\hspace{1em} Traditional stereo matching~\cite{zhang2009cross, sun2003stereo, hirschmuller2007stereo} typically consists of four steps: matching cost calculation, cost aggregation, disparity calculation, and disparity refinement. With the development of deep learning, each step in the traditional method can be replaced by a deep neural network for performance improvement. Zbontar and Lecun~\cite{zbontar2016stereo} first propose to replace the manually designed explicit metric function with the implicit metric learned by the network for computing matching cost. Seki et al.~\cite{seki2017sgm} propose SGM-Net which learns penalty parameters for semi-global matching (SGM).

%%%%%%%%%%%%%%%%%%%%%%%%%%%%%%%%%%%%%%%%%%%%%%%%%%%%%%%%%%%%%%%%%%%%%%%%
% 网络结构图
\begin{figure*}[ht]
    \centering
    \includegraphics[width=0.92\linewidth]{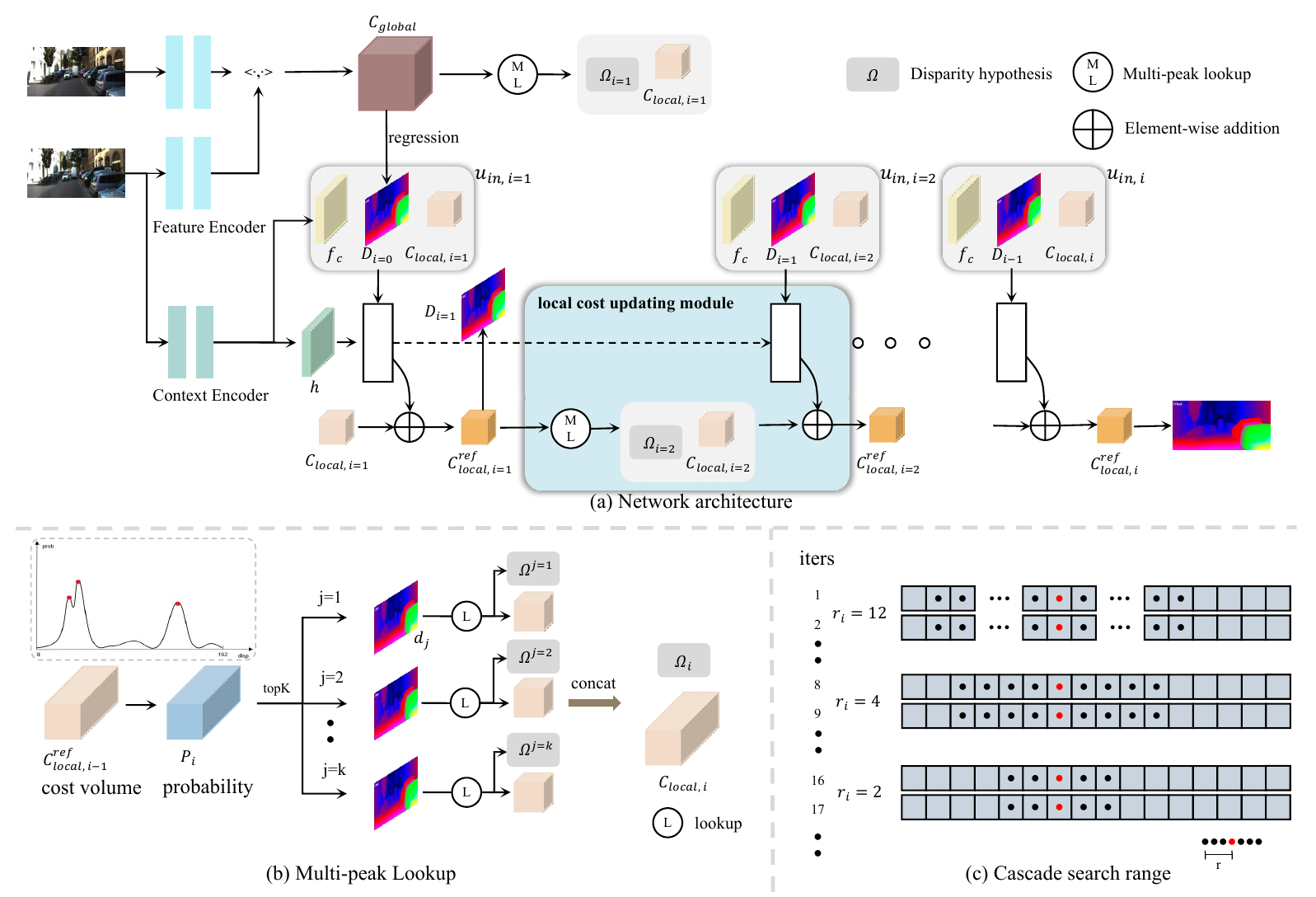}
    \vspace{0pt}
    \caption{Overview of MC-Stereo. (a) The architecture of MC-Stereo consists of three main components: feature extraction, cost volume construction, and iterative optimization. The core of iterative optimization is the local cost updating module that is based on multi-peak lookup and cascade search range. (b) Multi-peak lookup. Since the cost volume contains multiple peaks, we uniformly sample around top K disparities with the largest probability from the probability volume as our disparity hypothesis. (c) Cascade search range. We divide the search range into N levels and set different search ranges according to the number of iterations.}
    \label{fig:overview}
    \vspace{-10pt}
\end{figure*}
%%%%%%%%%%%%%%%%%%%%%%%%%%%%%%%%%%%%%%%%%%%%%%%%%%%%%%%%%%%%%%%%%%%%%%%%

In recent years, end-to-end networks have become the mainstream of stereo matching. Mayer et al.~\cite{mayer2016large} propose the first end-to-end disparity regression network Disp-Net which constructs a correlation volume based on CNN features and then aggregates the correlation volume using 2D convolutions. Finally, the disparity map is regressed from the aggregated volume. As the cost volume is the key affecting the final accuracy and efficiency of stereo matching, subsequent studies~\cite{guo2019group,xu2022attention,cheng2023coatrsnet,xu2023cgi,cheng2022region} mainly focus on cost volume construction and aggregation. In FastACV, Xu et al.~\cite{xu2022accurate} utilize a multi-peak strategy that selects the top-K disparities with the maximum probabilities to construct a sparse cost volume. Such a strategy can greatly reduce the computational cost without compromising accuracy excessively. Different from FastACV, in this work we not only select the top-K disparities but also uniformly samples around it to obtain more reliable distribution information.~\cite{yang2022non} and~\cite{mao2021uasnet} explore depth distribution and sampling strategies on a cascade architecture, respectively. In comparison, our approach focuses on iterative optimization architecture and is much more concise.

Although various cost volumes have been developed, these cost volumes could inevitably contain redundancy and noises arising from matching errors and ambiguities at ill-posed regions (e.g., occluded regions, textureless/transparent regions, etc.). To alleviate these problems, several CNN based cost aggregation methods have been developed. For instance, GCNet~\cite{kendall2017end} introduces 3D CNN based encoder-decoder to aggregate the cost volume. PSMNet~\cite{chang2018pyramid}, GWCNet~\cite{guo2019group} and ACVNet~\cite{xu2022attention} adopt a stacked hourglass block with 3D convolutional layers for cost aggregation. However, processing cost volumes using 3D convolutions is usually computationally expensive which limits its application to high-resolution cost volumes. GANet~\cite{zhang2019ga} develops a semi-global aggregation layer and a locally guided aggregation layer to replace 3D convolutions. However, such method increases the accuracy at the expense of the increased inference time. AANet~\cite{xu2020aanet} proposes an intra-scale and cross-scale cost aggregation algorithm to replace the traditional 3D convolution, which significantly improves the reasoning speed but leads to a decrease in accuracy. CoEx~\cite{bangunharcana2021correlate} introduces the Guided Cost volume Excitation (GCE) and demonstrates that the straightforward excitation of cost volume channels, guided by the image, can significantly enhance performance. Moreover, significant efforts~\cite{zhang2020domain, cai2020matching, zhang2022revisiting, liu2022graftnet} have been directed towards enhancing the generalizability of stereo matching networks.

Recently, the recurrent iterative optimization framework (e.g., RAFT-Stereo~\cite{lipson2021raft}), which directly optimizes the disparity maps via recurrent GRU updaters, has achieved impressive performance. Subsequent work CREStereo~\cite{li2022practical} and IGEV-Stereo~\cite{xu2023iterative} continue this framework. But they all have two problems: (1) Using single-peak lookup is not conducive to dealing with multi-peak distribution. (2) Using a single fixed search range limits the final convergence accuracy. In this work, we propose the multi-peak lookup strategy and cascade search range to address the above two issues.

\section{Approach}

\subsection{Network Architecture}

\hspace{1em} Figure~\ref{fig:overview} illustrates the architecture of our MC-Stereo which consists of three main modules: 1) feature extraction, 2) cost volume construction, and 3) iterative optimization. 

\noindent{\bf Feature Extraction.} We utilize two separate encoders, i.e., the \textit{feature encoder} and the \textit{context encoder}. For the \textit{feature encoder}, we use the first two stages of the ConvNeXt~\cite{liu2022convnet} pretrained on ImageNet~\cite{deng2009imagenet} to extract features for both the left and right images at 1/4 and 1/8 resolution. Then we fuse features of two different scales through a U-Net style up-sampling module to obtain the final feature maps with 1/4 resolution from the left image and right image respectively, denoted as $f_{l},f_{r} \in R^{C\times H\times W}$, where $C$=256, $H$ and $W$ are 1/4 of original image width and height. We use the same network architecture as the \textit{feature encoder} for the \textit{context encoder}, with the only difference being that the \textit{context encoder} is exclusively applied to the left image to obtain the context feature $f_{c}$ and the initial hidden state $h$ for the update operator.

\noindent{\bf Cost Volume Construction.} We construct a correlation volume based on left and right feature maps $f_{l},f_{r} \in R^{C\times H\times W}$ as:
\begin{equation}
{C_{init}}({d}, {x}, {y})=\left\langle f_{l}(x, y), f_{r}(x-d, y)\right\rangle
\end{equation}
where $\left \langle \cdot,\cdot \right \rangle$ is the inner product, $d$ is the disparity $\in [0,D_{max}/4]$, $D_{max}$ is the maximum disparity hypothesis (192), ($x$,$y$) represents the pixel coordinates. And the constructed cost volume $C_{init} \in R^{D_{max}/4\times H\times W}$ provides matching similarities between each pixel and all corresponding pixels in the right image with different disparities. Then, we apply average pooling along the disparity dimension of $C_{init}$ using a kernel size of 1,2 and equivalent stride, resulting in a 2-layer cost volume pyramid $\left \{ C^{1},C^{2}  \right \}$, which provides information of different receptive fields. For ease of description, we define the cost volume pyramid as $C_{global}$. When we perform a subsequent look-up operation on $C_{global}$, it actually retrieves each cost volume within the cost volume pyramid. During later iterative optimization, $C_{global}$ is not updated.

\noindent{\bf Iterative Optimization.} We convert $C_{init}$ into a probability volume via the softmax operation along the disparity dimension. Next, we use multi-peak lookup to select the top K disparities with the largest probability from the probability volume which serves as the initial disparity index for local cost volume aggregation from $C_{global}$. The updater produces a residual that updates the local cost volume, from which the disparity is computed. The top K disparity hypotheses with the highest probability are selected again from the updated local cost for the next iteration. We present details of iterative optimization with multi-peak lookup and cascade search range in Sec.~\ref{subsec:Updata Local Cost Volume}-\ref{subsec:Cascade Search Range}.

%-------------------------------------------------------------------------
\subsection{Iterative Optimization}

\label{subsec:Updata Local Cost Volume}
Preliminary: We obtain the final disparity by iterating the \textbf{local cost updating module}. In general, in the $ith$ iteration, the input of the local cost updating module
is the output of the updating module in previous iteration, i.e., the refined local cost $C_{local, i-1}^{ref}$ and the disparity map $D_{i-1}$ in $i$-1 iteration. Naturally, the outputs of the current iteration are denoted as $C_{local, i}^{ref}$ and $D_{i}$.

\textbf{local cost updating module.}
To be specific, for the $ith$ iteration, we first take the refined local cost volume (denoted as $C_{local, i-1}^{ref}$) which is refined in the previous iteration $i$-1 as the input of the multi-peak lookup operation (denoted as $ML$) to determine the disparity hypotheses set of the current iteration (denoted as $\Omega_{i}$) and construct the current local cost $C_{local, i}$ as Equation~\ref{ml}, the details are explained in Sec.~\ref{subsec:Multi-peak Lookup}. Then we utilize GRUs for updating the local cost volume. We concatenate the current iteration local cost $C_{local, i}$, previous iteration disparity map $D_{i -1}$ and the context feature $f_{c}$ as the input of GRUs (denoted as $u_{in,i}$) to obtain the residual of the local cost, denoted as
$\bigtriangleup C_{local, i}$.
$\bigtriangleup C_{local, i}$ is added with the current iteration local cost volume to obtain the refined local cost volume $C_{local, i}^{ref}$ as follow:
\begin{equation}
\Omega_{i}, C_{local, i} = ML(C_{local, i-1}^{ref})
\label{ml}
\end{equation}
\begin{equation}
\Delta C_{local, i}=G R U\left(u_{in,i}\right)
\end{equation}
\begin{equation}
C_{local, i }^{ref}=C_{local, i}+\Delta C_{local, i}
\end{equation}
% The update module aggregates the local cost volume to obtain the regularized local cost volume, which is 
We transform the refined local cost volume $C_{local, i}^{ref}$ into a probability distribution by the softmax operation. The current iteration disparity map $D_{i}$ can be obtained as:
\begin{equation}
D_{i}=\sum_{m\in \Omega_{i}} m \cdot \operatorname{softmax}\left(C_{{local, i}}^{ref}\right)
\label{eq4}
\end{equation}
In special cases, the first iteration, i.e., $i$=1, the input of the local updating module is from $C_{global}$ as shown in Figure \ref{fig:overview}.

%%%%%%%%%%%%%%%%%%%%%%%%%%%%%%%%%%%%%%%%%%%%%%%%%%%%%%%%%%%%%%%%%%%%%%%%
% ETH3D可视化图
\begin{figure*}[ht]
    \centering
    \includegraphics[width=0.9\linewidth]{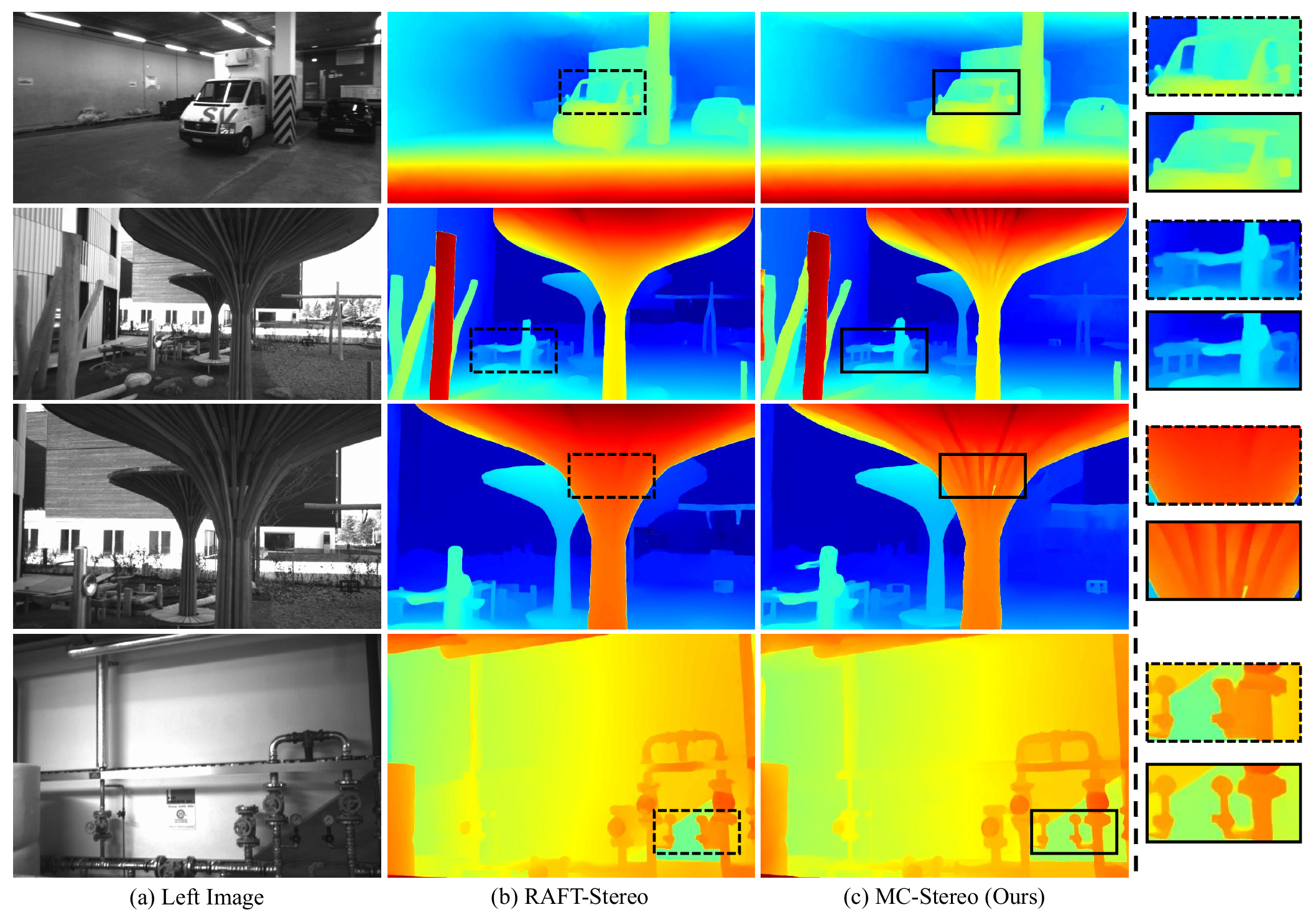}
    \vspace{-5pt}
    \caption{Qualitative results on ETH3D. The second and third columns are the results of RAFT-Stereo~\cite{lipson2021raft} and our MC-Stereo respectively.}
    \label{fig:vis_eth3d}
    \vspace{-10pt}
\end{figure*}
%%%%%%%%%%%%%%%%%%%%%%%%%%%%%%%%%%%%%%%%%%%%%%%%%%%%%%%%%%%%%%%%%%%%%%%%

%-------------------------------------------------------------------------
\subsection{Multi-Peak Lookup}
\label{subsec:Multi-peak Lookup}

\hspace{1em} One of the most important steps in recurrent optimization is to index the local cost volume from $C_{global}$. This lookup operation is represented by $L_{C}$ in RAFT-Stereo~\cite{lipson2021raft}. In the previous approach, only one single disparity obtained from the previous iteration is used to index the local cost volume.  However, we believe that such single-peak search strategy may be inadequate when dealing with multi-peak distribution. For disparity probabilities with a multi-peak distribution, the single-peak lookup strategy fails to sample the local cost volume in the appropriate range, resulting in the final disparity estimation falling into a wrong peak. To solve this problem, we propose the multi-peak lookup strategy. 

The process of multi-peak lookup is shown in Figure \ref{fig:overview} (b). 
To be specific, we first transform the previous iteration refined local cost volume $C_{local, i-1}^{ref}$ into a probability distribution $P_{i}$ through softmax, and the dimension of $P_{i}$ is $[N_{i-1},H/4,W/4]$, where $N_{i-1}$ is the number of sampling points in the $i$-1 iteration. In each iterations, $N_{i-1}=k\times (2r_{i-1}+1)$, where $k$ is the peak number of multi-peak look up, and $r_{i-1}$ is the search radius of the $i$-1 iteration. After obtaining the probability distribution, we obtain the $k$ disparity values $d_{j} \left ( j=1,\dots ,k \right )$ with the largest probabilities. For each disparity $d_{j}$, we uniformly sample around $d_{j}$ in the range of -$r$ and +$r$ to obtain the subset of the disparity hypotheses, denoted as $\Omega^{j}$, then merge all subsets together to obtain the final current iteration disparity hypotheses $\Omega_{i}$ as follow:
% we index the corresponding local cost volume from $C_{global}$, and save the corresponding disparity of each local cost volume in the list $R$:
\begin{equation}
{\Omega^{j}}=\left[d_{j}-{r}_{i}, \ldots, d_{j}, \ldots, d_{j}+{r}_{i}\right], (j = 1,\ldots,k)
\end{equation}
\begin{equation}
{\Omega_{i}}=\left[\Omega^{1}, \Omega^{2}, \ldots, \Omega^{k}\right]
\end{equation}

And we construct the current iteration local cost volume $C_{local, i}$ by indexing from the $C_{global}$ according to the depth hypotheses $\Omega_{i}$.

%%%%%%%%%%%%%%%%%%%%%%------------------------------------%%%%%%%%%%%%%%%%%%%%%%%%
\begin{table*}[t]
\centering
\setlength{\tabcolsep}{3.5mm}
\renewcommand{\arraystretch}{1.1}
\begin{tabular}{lcccccclccc}
\toprule[1.5pt]
\multirow{2}{*}{Method} & \multicolumn{4}{c}{All} &  & \multicolumn{2}{c}{Ref} & \multirow{2}{*}{\begin{tabular}[c]{@{}c@{}}Time\\ (s)\end{tabular}} \\ \cline{2-5} \cline{7-8}
 & 2-noc & 2-all & 3-noc & 3-all &  & 2-noc & 3-noc &  \\ \hline
PSMNet~\cite{chang2018pyramid} & 2.44 & 3.01 & 1.49 & 1.89 &  & 13.77 & 8.36 & 0.31 \\
GwcNet~\cite{guo2019group} & 2.16 & 2.71 & 1.32 & 1.70 &  & 12.49 & 7.80 & 0.20 \\
HITNet~\cite{tankovich2021hitnet} & 2.00 & 2.65 & 1.41 & 1.89 &  & 9.75 & 5.91 & 0.02 \\
CFNet~\cite{shen2021cfnet} & 1.90 & 2.43 & 1.23 & 1.58 &  & 9.91 & 5.96 & 0.18 \\
LEAStereo~\cite{cheng2020hierarchical} & 1.90 & 2.39 & 1.13 & 1.45 &  & 9.66 & 5.35 & 0.30 \\
ACVNet~\cite{xu2022attention} & 1.83 & 2.35 & 1.13 & 1.47 &  & 11.42 & 7.03 & 0.20 \\
LaC+GANet~\cite{liu2022local} & 1.72 & 2.26 & {\ul 1.05} & 1.42 &  & 10.40 & 6.02 & 1.60 \\
PCWNet~\cite{shen2022pcw} & {\ul 1.69} & 2.18 & \textbf{1.04} & {\ul 1.37} &  & 8.94 & 4.99 & 0.44 \\
IGEV-Stereo~\cite{xu2023iterative} & 1.71 & {\ul 2.17} & 1.12 & 1.43 & \multicolumn{1}{c}{} & {\ul 7.29} & {\ul 4.11} & 0.32 \\
CREStereo~\cite{li2022practical} & 1.72 & 2.18 & 1.14 & 1.46 & \multicolumn{1}{c}{} & 9.71 & 6.27 & 0.41 \\ \hline
MC-Stereo(Ours) & \textbf{1.55} & \textbf{1.99} & \textbf{1.04} & \textbf{1.34} &  & \textbf{6.82} & \textbf{4.10} & 0.40 \\ \hline
\end{tabular}
\setlength{\abovecaptionskip}{5pt}    
\setlength{\belowcaptionskip}{-10pt}
\caption{\textbf{Quantitative evaluation KITTI-2012~\cite{geiger2012we} benchmark.} Percentages of erroneous pixels for both non-occluded (noc) and all pixels are reported. ‘All’ denotes all pixels of the image, and ‘Ref’ denotes pixels of reflective area. Bold: Best, Underscore: Second best. The inference time is measured on a single NVIDIA RTX 3090 GPU at KITTI resolution (376×1248).}
\label{tab:stereo_kitti_12}
\end{table*}
%%%%%%%%%%%%%%%%%%%%%%------------------------------------%%%%%%%%%%%%%%%%%%%%%%%%

%-------------------------------------------------------------------------
\subsection{Cascade Search Range}

\label{subsec:Cascade Search Range}
\hspace{1em} The search range ($2\times {r}_{i}+1$) refers to the range of the grid for indexing from the global cost volume. The size of the search range affects the convergence speed and accuracy. The previous approaches use the same search range in all iterations, which could result in imprecision. At the beginning of recurrent optimization, applying a large search range can accelerate convergence and avoid falling into local optimum. In the later stage of optimization, a smaller search range can eliminate some disturbing items to improve matching accuracy. Based on this observation, we propose the cascade search range to further improve the matching accuracy. Our method searches with different ranges on the same resolution. Specifically, as shown in Figure \ref{fig:overview} (c), we divide the search range into three levels, i.e., 25, 9, and 5. Each level corresponds to particular GRU updaters. In the first few iterations, we use a large search range, and then reduce the search range step by step as follows:
\begin{equation}
r_{i}=\begin{cases}12,~i=1,\dots ,6
 \\4,~~~i=7,\dots ,16
 \\2,~~~i=17,\dots ,32

\end{cases}
\end{equation}

%-------------------------------------------------------------------------
\subsection{Loss Function}

\hspace{1em} We use the L1 loss between the predicted and ground truth disparity for a sequence of N refinement predictions of disparity, $\left \{ d_{1},\dots ,d_{N} \right \} $. Given ground truth $d_{gt}$, the loss is defined as:
\begin{equation}
L_{stereo}=\sum_{i=1}^{N} \gamma^{N-i}\left\|d_{g t}-d_{i}\right\|
\end{equation}
where $\gamma=0.9$.
% where $\gamma$ is the hyperparameter.

%-------------------------------------------------------------------------

\section{Experiments}
\label{sec:experiments}

\hspace{1em} This section details the experimental results of our proposed model on multiple datasets. We evaluate MC-Stereo on KITTI-2012~\cite{geiger2012we}, KITTI-2015~\cite{menze2015joint}, ETH3D~\cite{schops2017multi}, and Scene Flow~\cite{mayer2016large}. Our method achieves state-of-the-art performance on the KITTI-2012, KITTI-2015, and ETH3D leaderboards.

%-------------------------------------------------------------------------
\subsection{Implementation details}

\hspace{1em} We implement MC-Stereo in PyTorch~\cite{paszke2019pytorch} and use an AdamW~\cite{loshchilov2017decoupled} optimizer. And all experiments are performed on two NVIDIA RTX 3090 GPUs. We uniformly set the number of iterations to 32 when testing on KITTI-2012~\cite{geiger2012we}, KITTI-2015~\cite{menze2015joint}, ETH3D~\cite{schops2017multi} leaderboard. Data augmentation is used including saturation change, image perturbance, and random scales. Following previous work, we pretrain our model on Scene Flow~\cite{mayer2016large} training set for 200k steps with a batch size of 8. For the KITTI-2012 and KITTI-2015 leaderboards, we finetune the pre-trained Scene Flow model on the mixed KITTI-2012 and KITTI-2015 training set. For ETH3D, we finetune the pre-trained Scene Flow model on the mixed CREStereo Dataset~\cite{li2022practical}, InStereo2K~\cite{bao2020instereo2k} and ETH3D~\cite{schops2017multi} dataset.

%-------------------------------------------------------------------------
\subsection{MC-Stereo Performance}
\subsubsection{KITTI-2012}
\hspace{1em} We submit MC-Stereo to the KITTI-2012~\cite{geiger2012we} stereo benchmark. Among all published methods on the KITTI-2012 leaderboard, MC-Stereo ranks $1^{st}$ on several evaluation metrics. Details are shown in Table~\ref{tab:stereo_kitti_12}. On Out-Noc under 2 pixels error threshold, our MC-Stereo outperforms the next best method PCWNet~\cite{shen2022pcw} by 8.28\%. On Out-All under 2 pixels error threshold, our MC-Stereo surpasses the second best method IGEV-Stereo~\cite{xu2023iterative} by 8.29\%. On Out-Noc under 2 pixels error threshold in reflective area, our MC-Stereo outperforms the next best method IGEV-Stereo~\cite{xu2023iterative} by 6.44\%.

\subsubsection{KITTI-2015}
\hspace{1em} At the time of writing this paper, our MC-Stereo ranks $1^{st}$ on the KITTI-2015~\cite{menze2015joint} leaderboard among all published methods (See Table~\ref{tab:stereo_kitti_15}). On the percentage of erroneous foreground pixels, our MC-Stereo outperforms RAFT-Stereo~\cite{lipson2021raft} and IGEV-Stereo~\cite{xu2023iterative} by 13.15\% and 5.99\%. Figure~\ref{fig:vis_kitti} shows qualitative results on KITTI. Our MC-Stereo performs better in reflective areas.

%%%%%%%%%%%%%%%%%%%%%%------------------------------------%%%%%%%%%%%%%%%%%%%%%%%%
\begin{table*}[]
\centering
\setlength{\tabcolsep}{3.0mm}
\renewcommand{\arraystretch}{1.1}
\begin{tabular}{lccclcccc}
\toprule[1.5pt]
\multirow{2}{*}{Method} & \multicolumn{3}{c}{All} &  & \multicolumn{3}{c}{Noc} & \multirow{2}{*}{\begin{tabular}[c]{@{}c@{}}Time\\ (s)\end{tabular}} \\ \cline{2-4} \cline{6-8}
 & D1-bg & D1-fg & D1-all &  & D1-bg & D1-fg & D1-all &  \\ \hline
PSMNet~\cite{chang2018pyramid} & 1.86 & 4.62 & 2.32 &  & 1.71 & 4.31 & 2.14 & 0.31 \\
GwcNet~\cite{guo2019group} & 1.74 & 3.93 & 2.11 &  & 1.61 & 3.49 & 1.92 & 0.20 \\
HITNet~\cite{tankovich2021hitnet} & 1.74 & 3.2 & 1.98 &  & 1.54 & 2.72 & 1.74 & 0.02 \\
CFNet~\cite{shen2021cfnet} & 1.54 & 3.56 & 1.88 &  & 1.43 & 3.25 & 1.73 & 0.18 \\
LEAStereo~\cite{cheng2020hierarchical} & 1.40 & 2.91 & 1.65 &  & 1.29 & 2.65 & 1.51 & 0.30 \\
ACVNet~\cite{xu2022attention} & {\ul 1.37} & 3.07 & 1.65 &  & {\ul 1.26} & 2.84 & 1.52 & 0.20 \\
LaC+GANet~\cite{liu2022local} & 1.44 & 2.83 & 1.67 &  & {\ul 1.26} & 2.64 & {\ul 1.49} & 1.60 \\
PCWNet~\cite{shen2022pcw} & {\ul 1.37} & 3.16 & 1.67 &  & {\ul 1.26} & 2.93 & 1.53 & 0.44 \\
RAFT-Stereo~\cite{lipson2021raft} & 1.75 & 2.89 & 1.91 &  & - & - & - & 0.38 \\
IGEV-Stereo~\cite{xu2023iterative} & 1.38 & {\ul 2.67} & {\ul 1.59} & \multicolumn{1}{c}{} & 1.27 & 2.62 & 1.49 & 0.32 \\
CREStereo~\cite{li2022practical} & 1.45 & 2.86 & 1.69 & \multicolumn{1}{c}{} & 1.33 & {\ul 2.60} & 1.54 & 0.41 \\ \hline
MC-Stereo(Ours) & \textbf{1.36} & \textbf{2.51} & \textbf{1.55} &  & \textbf{1.24} & \textbf{2.55} & \textbf{1.46} & 0.40 \\ \hline
\end{tabular}
\setlength{\abovecaptionskip}{5pt}    
\setlength{\belowcaptionskip}{-10pt}
\caption{\textbf{Quantitative evaluation KITTI-2015~\cite{menze2015joint} benchmark.} Percentages of disparity outliers D1 for background, foreground, and all pixels are reported. ‘All’ denotes all pixels of the image, and ‘Noc’ denotes the non-occluded pixels. Bold: Best, Underscore: Second best.}
\label{tab:stereo_kitti_15}
\end{table*}
%%%%%%%%%%%%%%%%%%%%%%------------------------------------%%%%%%%%%%%%%%%%%%%%%%%%

\subsubsection{ETH3D}
\hspace{1em} Table~\ref{tab:stereo_eth3d} summarizes the results on the ETH3D~\cite{schops2017multi} leaderboard. MC-Stereo achieves state-of-the-art performance on ETH3D. At the time of writing this paper, MC-Stereo ranks 1st on the 50\% error quantile. Figure~\ref{fig:vis_eth3d} shows qualitative results on ETH3D.

\subsubsection{Scene Flow}
\begin{figure*}[ht]
    \centering
    \includegraphics[width=0.9\linewidth]{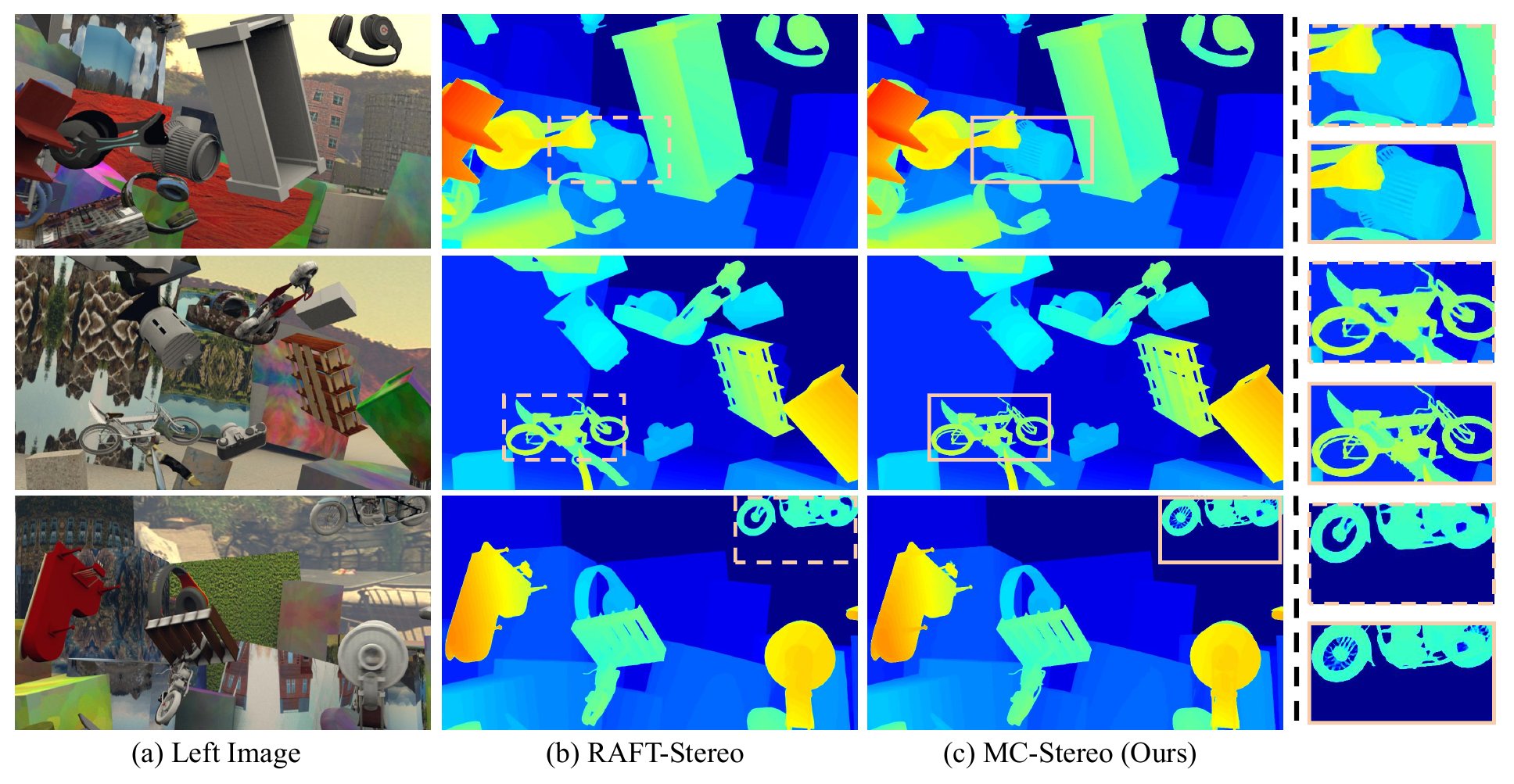}
    \vspace{-5pt}
    \caption{Qualitative results on Scene Flow. The second and third columns are the results of RAFT-Stereo~\cite{lipson2021raft} and our MC-Stereo respectively. Our MC-Stereo effectively captures intricate details in objects that have fine structures.}
    \label{fig:vis_sceneflow}
    \vspace{-10pt}
\end{figure*}

\hspace{1em} On Scene Flow~\cite{mayer2016large} test set, MC-Stereo achieves competitive results. Details results are summarized in Table~\ref{tab:stereo_sceneflow}. Visual comparisons are shown in Figure~\ref{fig:vis_sceneflow}. Our MC-Stereo effectively captures intricate details in objects that have fine structures.

%%%%%%%%%%%%%%%%%%%%%%%%%%%%%%%%%%%%%%%%%%%%%%%%%%%%%%%%%%%%%%%%%%%%%%%%
% Sceneflow可视化图
%%%%%%%%%%%%%%%%%%%%%%%%%%%%%%%%%%%%%%%%%%%%%%%%%%%%%%%%%%%%%%%%%%%%%%%%

%%%%%%%%%%%%%%%%%%%%%%------------------------------------%%%%%%%%%%%%%%%%%%%%%%%%
\begin{table}[]
\centering
\setlength{\tabcolsep}{1.8mm}
\renewcommand{\arraystretch}{1.2}
\begin{tabular}{lcccc}
\toprule[1.5pt]
Method & bad 4.0(\%) & AvgErr & \begin{tabular}[c]{@{}c@{}}50\% error\\ quantile\end{tabular} \\ \hline
HITNet~\cite{tankovich2021hitnet} & 0.19 & 0.20 & 0.10 \\
GwcNet~\cite{guo2019group} & 0.50 & 0.35 & 0.20 \\
CFNet~\cite{shen2021cfnet} & 0.56 & 0.27 & 0.15 \\
ACVNet~\cite{xu2022attention} & 0.20 & 0.23 & 0.15 \\
RAFT-Stereo~\cite{lipson2021raft} & 0.15 & 0.18 & 0.10 \\
CREStereo~\cite{li2022practical} & \textbf{0.10} & \textbf{0.13} & {\ul 0.09} \\
IGEV-Stereo~\cite{xu2023iterative} & {\ul 0.11} & {\ul 0.14} & {\ul 0.09} \\ \hline
MC-Stereo(Ours) & \textbf{0.10} & {\ul 0.14} & \textbf{0.08} \\ \hline
\end{tabular}
\setlength{\abovecaptionskip}{5pt}    
\setlength{\belowcaptionskip}{-10pt}
\caption{\textbf{Quantitative evaluation on ETH3D~\cite{schops2017multi} benchmark.} Fraction of pixels with errors larger than 4 (bad 4.0), the per-pixel average disparity error, and 50\% error quantile 
on non-occluded area are reported. Bold: Best, Underscore: Second best.}
\label{tab:stereo_eth3d}
\end{table}
%%%%%%%%%%%%%%%%%%%%%%------------------------------------%%%%%%%%%%%%%%%%%%%%%%%%

%%%%%%%%%%%%%%%%%%%%%%------------------------------------%%%%%%%%%%%%%%%%%%%%%%%%
\begin{table}[]
\centering
\setlength{\tabcolsep}{2.0mm}
\renewcommand{\arraystretch}{1.2}
\begin{tabular}{lccc}
\toprule[1.5pt]
\multirow{2}{*}{Method} & \multicolumn{3}{c}{Scene Flow} \\ \cline{2-4} 
 & EPE & \textgreater{}1px(\%) & \textgreater{}3px(\%) \\ \hline
PSMNet~\cite{chang2018pyramid} & 1.07 & 10.90 & 4.40 \\
GwcNet~\cite{guo2019group} & 0.79 & 8.19 & 3.40 \\
ACV-Net~\cite{xu2022attention} & 0.48 & - & - \\
LEAStereo~\cite{cheng2020hierarchical} & 0.78 & 7.82 & - \\
RAFT-Stereo~\cite{lipson2021raft} & 0.60 & 6.77 & 3.18 \\
IGEV-Stereo~\cite{xu2023iterative} & {\ul 0.47} & {\ul 5.21} & {\ul 2.48} \\ \hline
MC-Stereo(Ours) & \textbf{0.45} & \textbf{4.97} & \textbf{2.32} \\ \hline
\end{tabular}
\setlength{\abovecaptionskip}{5pt}    
\setlength{\belowcaptionskip}{-10pt}
\caption{\textbf{Quantitative evaluation on Scene Flow~\cite{mayer2016large} test set.} Percentages of erroneous pixels and average end-point errors are reported. Bold: Best, Underscore: Second best.}
\label{tab:stereo_sceneflow}
\end{table}
%%%%%%%%%%%%%%%%%%%%%%------------------------------------%%%%%%%%%%%%%%%%%%%%%%%%

%-------------------------------------------------------------------------
\subsection{Ablation Study}

\hspace{1em} A series of ablation studies are conducted to verify the effectiveness of each component in MC-Stereo. All ablation experiments are trained on the Scene Flow~\cite{mayer2016large} dataset for 200k iterations with a batch size of 8.

\noindent{\bf Cascade Search Range:} We conduct a comprehensive ablation experiment on the cascade search range, and the results are summarized in Table~\ref{tab:ablation_cascade}. The search radius of '4' is commonly employed in existing iterative optimization frameworks. With a total of 32 iterations, we divide the iterative optimization process into three stages. Initially, we expand the search radius in the first stage to conduct the experiment, subsequently narrowing the search range in the third stage based on this initial analysis. The results indicate that the cascade search range, based on the coarse-to-fine approach, outperforms a single fixed search range.

\noindent{\bf Multi-Peak Lookup:} Table~\ref{tab:ablation_ml} presents the results of the ablation experiment on multi-peak lookup. The label 'No' signifies the utilization of the single-peak lookup strategy, while the variable 'K' represents the number of peaks set in the multi-peak lookup approach. The findings demonstrate a consistent performance improvement between multi-peak and single-peak lookup strategies. Moreover, as K approaches 3, the benefits gained from employing the multi-peak lookup strategy tend to plateau.

%%%%%%%%%%%%%%%%%%%%%%------------------------------------%%%%%%%%%%%%%%%%%%%%%%%%
\begin{table}[]
\centering
\setlength{\tabcolsep}{1.0mm}
\renewcommand{\arraystretch}{1.2}
\begin{tabular}{ccccccc}
\toprule[1.5pt]
\multirow{2}{*}{} & \multirow{2}{*}{\begin{tabular}[c]{@{}c@{}}Iter\\ 1$\sim$6\end{tabular}} & \multirow{2}{*}{\begin{tabular}[c]{@{}c@{}}Iter\\ 7$\sim$16\end{tabular}} & \multirow{2}{*}{\begin{tabular}[c]{@{}c@{}}Iter\\ 17$\sim$32\end{tabular}} & \multicolumn{3}{c}{Scene Flow} \\ \cline{5-7} 
 &  &  &  & EPE & \textgreater{}1px(\%) & \textgreater{}3px(\%) \\ \hline
\multirow{3}{*}{\begin{tabular}[c]{@{}c@{}}Search\\ Radius\end{tabular}} & 4 & 4 & 4 & 0.53 & 5.96 & 2.73 \\
 & 12 & 4 & 4 & 0.51 & 5.80 & 2.63 \\
 & 12 & 4 & 2 & 0.50 & 5.74 & 2.62 \\ \hline
\end{tabular}
\setlength{\abovecaptionskip}{5pt}    
\setlength{\belowcaptionskip}{-10pt}
\caption{\textbf{Ablation study of cascade search range.} }
\label{tab:ablation_cascade}
\end{table}
%%%%%%%%%%%%%%%%%%%%%%------------------------------------%%%%%%%%%%%%%%%%%%%%%%%%

\noindent{\bf Pretrained Feature Extractor:} Accurate feature representation learning is crucial for successful learn-based stereo matching. In Table~\ref{tab:ablation_fe}, we compare the performance of a feature extraction network initialized with random parameters to one loaded with pre-trained parameters on ImageNet~\cite{deng2009imagenet}. The results demonstrate that pre-training on ImageNet enhances the characterization ability of the extracted features, thereby benefiting stereo matching. It is worth noting that pre-training on ImageNet is primarily intended for classifying tasks. However, the dissimilarities between matching tasks may restrict the complete benefits of pre-training. Exploring dedicated pre-training methods for matching tasks is a worthwhile endeavor for future research.

%-------------------------------------------------------------------------

\section{Limitation}
\label{sec:limitation}

\hspace{1em} Our multi-peak lookup strategy and cascade search range achieve significant performance improvements, but the hyperparameters (such as the number of peaks for multi-peak lookup, series division, search range, etc.) are set empirically. When dealing with new scenarios, these parameters may need to be readjusted for best performance. In future work, we will focus on making hyperparameter settings dynamic, allowing the network to choose independently according to the scenario.

%%%%%%%%%%%%%%%%%%%%%%------------------------------------%%%%%%%%%%%%%%%%%%%%%%%%
\begin{table}[t]
\centering
\setlength{\tabcolsep}{0.9mm}
\renewcommand{\arraystretch}{1.1}
\begin{tabular}{cccccc}
\toprule[1.5pt]
\multirow{2}{*}{} & \multirow{2}{*}{Variations} & \multicolumn{3}{c}{Scene Flow} & \multirow{2}{*}{\begin{tabular}[c]{@{}c@{}}Param\\ (M)\end{tabular}} \\ \cline{3-5}
 &  & EPE & \textgreater{}1px(\%) & \textgreater{}3px(\%) &  \\ \hline
\multirow{4}{*}{\begin{tabular}[c]{@{}c@{}}Multi-peak\\ Lookup\end{tabular}} & No & 0.50 & 5.74 & 2.62 & 21.2M \\
 & K=2 & 0.49 & 5.43 & 2.56 & 21.3M \\
 & {\ul K=3} & 0.48 & 5.38 & 2.55 & 21.4M \\
 & K=4 & 0.48 & 5.37 & 2.55 & 21.5M \\ \hline
\end{tabular}
\setlength{\abovecaptionskip}{5pt}    
\setlength{\belowcaptionskip}{-10pt}
\caption{\textbf{Ablation study of multi-peak lookup.} Settings used in our final model are underlined.}
\label{tab:ablation_ml}
\end{table}
%%%%%%%%%%%%%%%%%%%%%%------------------------------------%%%%%%%%%%%%%%%%%%%%%%%%

%%%%%%%%%%%%%%%%%%%%%%------------------------------------%%%%%%%%%%%%%%%%%%%%%%%%
\begin{table}[t]
\centering
\setlength{\tabcolsep}{0.9mm}
\renewcommand{\arraystretch}{1.1}
\begin{tabular}{ccccc}
\toprule[1.5pt]
\multirow{2}{*}{Experiment} & \multirow{2}{*}{Variations} & \multicolumn{3}{c}{Scene Flow} \\ \cline{3-5} 
 &  & EPE & \textgreater{}1px(\%) & \textgreater{}3px(\%) \\ \hline
\multirow{2}{*}{\begin{tabular}[c]{@{}c@{}}Pretrained\\ Feature Extractor\end{tabular}} &  & 0.48 & 5.38 & 2.55 \\
 & $\surd$ & 0.45 & 4.97 & 2.32 \\ \hline
\end{tabular}
\setlength{\abovecaptionskip}{5pt}    
\setlength{\belowcaptionskip}{-10pt}
\caption{\textbf{Ablation study of feature extractor.} }
\label{tab:ablation_fe}
\end{table}
%%%%%%%%%%%%%%%%%%%%%%------------------------------------%%%%%%%%%%%%%%%%%%%%%%%%

\section{Conclusion}
\label{sec:conclusion}

\hspace{1em} We have proposed MC-Stereo, a new learning based method for Stereo Matching. The multi-peak lookup strategy improves the ability of the model to deal with multi-peak distribution. Cascade search range combines the idea of coarse to fine into the iterative optimization framework. Our method ranks first among all published methods on the KITTI-2012 and KITTI-2015 leaderboards, and also achieves state-of-the-art performance on the ETH3D benchmark.

\noindent\textbf{Acknowledgment.} This work was supported in part by the National Natural Science Foundation of China under Grants 62122029, 62061160490, and U20B2064.

{
    \small
    \bibliographystyle{ieeenat_fullname}
    \bibliography{main}
}

\end{document}